\documentclass[letterpaper, 10 pt, conference]{ieeeconf}  

\usepackage{amsmath}
\usepackage{multirow}
\usepackage{graphicx}
\usepackage{caption}
\usepackage{amssymb}
\usepackage{float}
\usepackage[pagebackref=true,breaklinks=true,letterpaper=true,colorlinks,bookmarks=false,urlcolor=red,citecolor=cyan]{hyperref}
\usepackage{cleveref}
\usepackage{xcolor} 
\usepackage{placeins} 
\usepackage{varwidth}
\usepackage{changepage} 
\usepackage{caption} 

\IEEEoverridecommandlockouts                              

\title{\LARGE \bf
TTTFusion: A Test-Time Training-Based Strategy for Multimodal Medical Image Fusion in Surgical Robots
}

\author{Qinhua Xie$^{1}$, Hao Tang$^{2*}$
\thanks{$^{1}$School of Data Science and Engineering, East China Normal University,
        Shanghai, China
        {\tt\small 10214102413@stu.ecnu.edu.cn}}%
\thanks{$^{2}$School of Computer Science, Peking University, Beijing, China {\tt\small haotang@pku.edu.cn}}%
\thanks{*Corresponding author}
}

\begin{document}

\maketitle
\thispagestyle{empty}
\pagestyle{empty}

\begin{abstract}


With the increasing use of surgical robots in clinical practice, enhancing their ability to process multimodal medical images has become a key research challenge. Although traditional medical image fusion methods have made progress in improving fusion accuracy, they still face significant challenges in real-time performance, fine-grained feature extraction, and edge preservation.
In this paper, we introduce TTTFusion, a Test-Time Training (TTT)-based image fusion strategy that dynamically adjusts model parameters during inference to efficiently fuse multimodal medical images. By adapting the model during the test phase, our method optimizes the parameters based on the input image data, leading to improved accuracy and better detail preservation in the fusion results.
Experimental results demonstrate that TTTFusion significantly enhances the fusion quality of multimodal images compared to traditional fusion methods, particularly in fine-grained feature extraction and edge preservation. This approach not only improves image fusion accuracy but also offers a novel technical solution for real-time image processing in surgical robots.

\end{abstract}

\section{INTRODUCTION}


Surgical robots are becoming increasingly prevalent in modern medicine, particularly in minimally invasive and complex surgeries, where their precision, stability, and flexibility have led to significant advances \cite{li2021clinical,zhu2021intelligent}. However, despite their great potential in clinical practice, optimizing their performance remains a challenge, especially in the fusion and processing of multimodal medical images. Surgical robots must integrate image data from different imaging modalities, such as CT and MRI, which vary significantly in spatio-temporal distribution, structural details, and physical properties. Efficiently fusing these multimodal images while preserving fine-grained pathological features is crucial to enhancing the performance of surgical robots.


In recent years, deep learning methods have been widely applied to medical image processing, particularly in image fusion tasks, achieving remarkable progress. Traditional fusion techniques, such as those based on Convolutional Neural Networks (CNNs) \cite{liu2017medical} and Transformers \cite{vs2022image}, have demonstrated their ability to combine information from different modalities effectively. However, these methods still face challenges in real-time applications, edge information preservation, and cross-modal feature alignment \cite{azam2022review}. Moreover, existing approaches typically assume that the models are fully trained in advance and lack the ability to dynamically adjust to real-time data during testing.

\begin{figure}[t]
    \centering
    \includegraphics[width=0.5\textwidth]{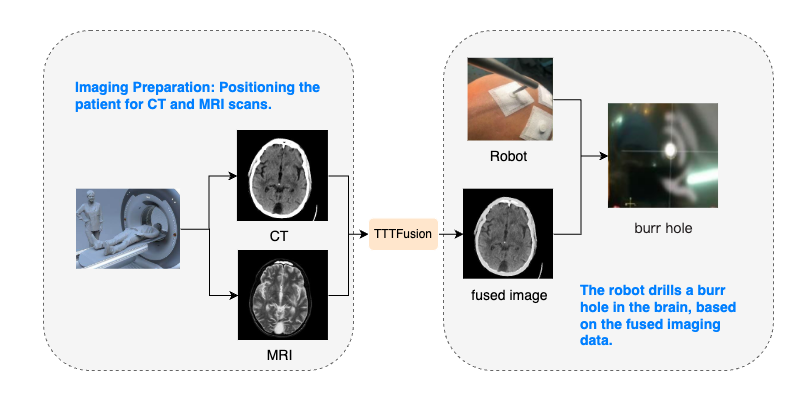}
    \caption{The figure shows the process of using a surgical robot for assistance in brain surgery. First, the patient is properly positioned for both CT and MRI imaging. Then, the robot drills a burr hole in the brain based on the fused images and performs the surgery precisely (robot image source Han et al. \cite{han2024efficacy}).
}
    \label{fig:robot}
\end{figure}

To address these challenges, this study introduces Test-Time Training Fusion (TTTFusion), a novel image fusion strategy that integrates Test-Time Training \cite{sun2024learning} with image fusion techniques. As illustrated in Figure \ref{fig:robot}, TTTFusion is applied in this way in the brain surgical robot. TTTFusion enables the model to dynamically adjust during inference, optimizing its parameters based on input multimodal image data for more precise fusion. Unlike traditional methods, which rely on static model parameters, TTTFusion incrementally adapts the model in real time, significantly improving fine-grained feature retention and edge information fusion.


The primary contribution of this paper is the introduction of the TTTFusion strategy, which enhances fine-grained feature extraction and edge preservation in multimodal medical images by allowing adaptive model adjustments during inference. Compared to existing fusion methods, TTTFusion dynamically fine-tunes model parameters based on the unique characteristics of the input images, improving fusion quality. This approach not only improves the accuracy of the image fusion, but also effectively addresses modality differences, significantly improving the visualization of complex lesions. Ultimately, TTTFusion presents a promising new pathway for real-time image processing in surgical robots, further advancing the development of surgical robotics technology.

\section{Related Work}

\subsection{Image Segmentation}


Early image segmentation methods primarily relied on techniques such as edge detection \cite{senthilkumar2018} and region growing \cite{preetha2012}. Although effective for simpler images, these methods struggled with complex structures and low-contrast medical images. With the emergence of deep learning, the U-Net architecture \cite{ronneberger2015u} has become a widely adopted approach to medical image segmentation. Its symmetric encoder-decoder structure and skip connections have significantly improved segmentation accuracy, particularly in handling complex medical images \cite{azad2024medical}. Furthermore, 3D U-Net \cite{cciccek20163d} extended the traditional U-Net model to three-dimensional data, enabling successful applications in volumetric imaging, such as CT \cite{zettler2021comparison} and MRI \cite{hwang20193d}.
As technology advances, image segmentation techniques have found widespread applications in the medical field, including tumor detection, organ segmentation, and brain imaging. The primary goal of these methods is to automatically extract regions of interest (ROIs) from medical images, providing clinicians with precise diagnostic support. Image segmentation not only aids in the diagnosis of tumors and other diseases but also plays a crucial role in treatment planning, surgical navigation, and radiotherapy.

In recent years, the introduction of new deep learning techniques has further enhanced the accuracy and efficiency of image segmentation. These advances have demonstrated great potential in multimodal image integration and real-time surgical applications, paving the way for more precise and efficient medical image analysis.

\subsection{Image Fusion}

Image fusion is the process of combining medical images from different modalities to provide more comprehensive information, which is particularly crucial for surgical robots and multimodal image analysis. Traditional multimodal fusion methods, such as wavelet transform \cite{singh2014} and Principal Component Analysis (PCA) \cite{he2010}, effectively extract key features from different modalities and preserve multi-scale information during fusion. However, they face challenges in real-time processing and fine-grained feature preservation.
With the rise of deep learning, convolutional neural network (CNN)-based fusion methods \cite{singh2019} have achieved significant advancements. For instance, SwinFusion \cite{ma2022} utilizes deep convolutional networks and multi-scale feature extraction to fuse images from different modalities, enhancing image accuracy and detail retention. Another deep learning-based approach, MRSCFusion \cite{xie2023}, improves edge preservation using multi-scale convolutional networks and edge-enhancing strategies, allowing for more precise fusion of complex medical images.
Furthermore, MambaFusion \cite{xie2024} integrates reinforcement learning with CNNs to dynamically adjust fusion strategies, further enhancing detail retention and excelling in complex fusion tasks. FusionGAN \cite{ma2019} applies Generative Adversarial Networks (GANs) to optimize fusion results, improving multimodal image integration. Meanwhile, DDcGAN \cite{ma2020} combines GANs with multi-resolution image fusion techniques, demonstrating strong performance in fusing low-resolution images.

Although these methods have significantly improved fusion accuracy and real-time processing, challenges remain in handling complex structures, image noise, and edge retention. This is particularly critical in surgical robots, where high computational efficiency and precision are essential for real-world applications.

\subsection{Integrated Applications in Surgical Robots}



\begin{figure*}[t]
    \centering
    \includegraphics[width=1\textwidth]{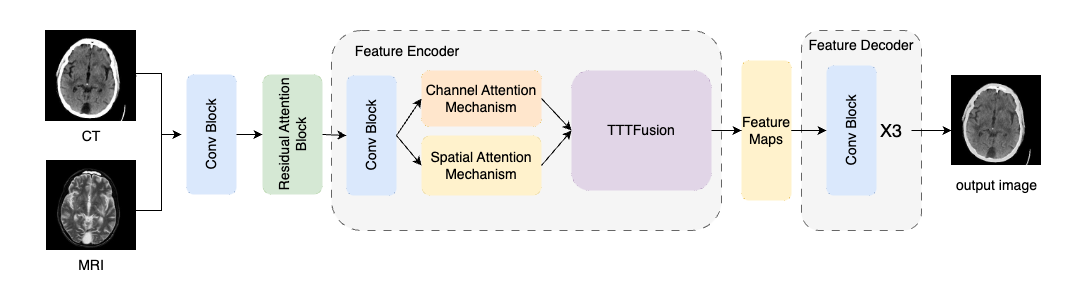}
    \caption{The application of the multimodal medical image fusion strategy based on Test-Time Training proposed by us in surgical robots.}
    \label{fig:all}
\end{figure*}

Robotic surgery has transformed modern surgical practices, particularly in minimally invasive procedures, by allowing greater precision and reducing patient recovery times. For example, the Da Vinci Surgical System has allowed surgeons to perform highly complex procedures with enhanced accuracy, especially in urology, by improving depth perception and movement flexibility through three-dimensional imaging \cite{rivero2023robotic}.
Despite these advancements, surgical robots still face significant challenges, particularly in automation and intelligence. In the future, the integration of AI is expected to play a key role in improving robotic surgery. Machine learning algorithms will enable real-time surgical process analysis, evaluate surgeon techniques, and optimize surgical planning, ultimately improving outcomes \cite{reddy2023advancements}.

Although robotic surgery has made remarkable progress, challenges such as high costs, equipment maintenance, and surgeon training continue to limit its widespread adoption. However, with ongoing advancements in machine learning, emerging technologies are expected to further enhance robotic surgery by improving operational precision and enabling real-time decision-making, thereby reducing complications \cite{knudsen2024clinical}. Currently, the combination of automation systems and virtual training platforms is becoming a major focus in surgical robotics. The integration of augmented reality (AR) and real-time data feedback is helping surgeons perform more complex procedures with greater accuracy and confidence \cite{bramhe2022robotic}.

\section{Methods}

\subsection{Architecture Overview} 


In this paper, we propose a multimodal medical image fusion method based on TTTFusion, which is designed to enhance the accuracy and quality of medical images by efficiently integrating features from diverse modalities, such as Computed Tomography (CT) and Magnetic Resonance Imaging (MRI). The proposed approach is clearly illustrated in Figure \ref{fig:all}, and it consists of three primary components: Feature Extraction, the TTTFusion Module, and Image Reconstruction.

For the Feature Extraction step, the process begins with the input of CT and MRI images. Firstly, these images are processed through a Conv Block (Convolutional Block). This block is responsible for performing initial convolutional operations on the input images to extract basic features. Subsequently, the output of the first Conv Block enters the Residual Attention Block. The Residual Attention Block not only helps in retaining the important features but also enhances them by adding attention mechanisms, which can focus on the key regions of the images. After that, the processed data goes through another Conv Block. At this stage, both low - level and high - level features are extracted from the input CT and MRI images using Convolutional Neural Networks (CNNs).

Next, these extracted features are fed into the TTTFusion Module. Inside this module, there are two important sub - mechanisms: the Channel Attention Mechanism and the Spatial Attention Mechanism. The Channel Attention Mechanism focuses on selectively emphasizing or suppressing different channels of the feature maps, based on the importance of the information carried by each channel. Meanwhile, the Spatial Attention Mechanism pays attention to different spatial locations in the feature maps, highlighting the regions that are more relevant for the fusion process. The TTTFusion Module then dynamically adjusts the feature fusion strategy according to the characteristics of the input images, ensuring an optimal combination of information from different imaging modalities.

Finally, in the Image Reconstruction step, the fused features, which are outputted as Feature Maps from the TTTFusion Module, are decoded back into the image space. This is achieved by passing the Feature Maps through a series of three Conv Blocks. The final output of these operations is the reconstructed and fused image, which combines complementary information from both CT and MRI modalities, providing a more comprehensive and accurate representation for medical applications. 

\begin{figure*}[t]
    \centering
    \includegraphics[width=0.9\textwidth]{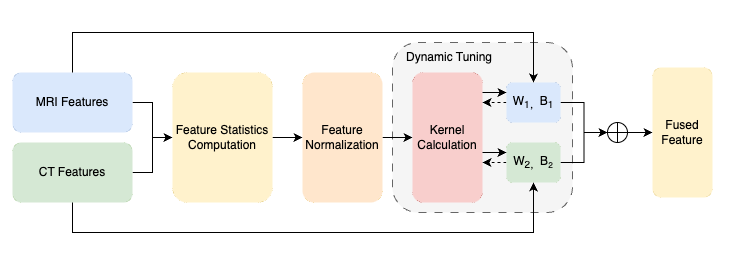}
    \caption{Detailed architecture of the proposed TTTFusion, which is a dynamic feature fusion framework for multimodal medical image processing.}
    \label{fig:TTTfusion2}
\end{figure*}

\subsection{The Proposed TTTFusion Module}

In the field of medical image fusion, traditional methods face significant limitations. Techniques such as weighted averaging and feature concatenation are simplistic and do not fully exploit the unique characteristics of different medical imaging modalities. For example, CT images excel at visualizing bone structures and dense tissues, while MRI scans provide detailed information about soft tissues and certain physiological processes. However, traditional methods rely on fixed fusion strategies that cannot adapt to the inherent variability and complexity among imaging modalities. As a result, these approaches often produce suboptimal fusion results, where critical diagnostic information may be overlooked or misrepresented, ultimately affecting the accuracy of medical diagnoses.

\textbf{TTT as a Solution.} TTT emerges as a powerful solution to these challenges by allowing dynamic model adjustments during inference. In the context of medical image fusion, this adaptability is crucial, as it allows the model to handle the diverse and often unpredictable nature of different medical modalities. Using TTT, the fusion process can be optimized for each individual input, ensuring that the distinct features of each modality are maximally utilized and integrated. This results in higher-quality fused images that provide more accurate and informative insights for medical professionals.

One of the key advantages of TTT is its ability to adapt to the specific characteristics of each input medical image, such as variations in image quality, patient anatomy, and imaging artifacts. By dynamically fine-tuning model parameters during inference, TTT significantly enhances the performance of medical image fusion, producing more reliable images for clinical decision-making.

\textbf{TTT vs. Traditional Static Training.} TTT represents a major shift from traditional static training methods. In conventional machine learning, models are trained on a fixed dataset with predefined hyperparameters and parameter updates based on the training data distribution. Once trained, the model’s behavior remains unchanged during inference, regardless of the specific characteristics of the input data. This static approach fails to account for the diversity and variability of real-world medical images, where significant differences exist between patients, imaging conditions, and disease presentations.

In contrast, TTT allows the model to adjust its parameters during the test phase, adapting to the unique features and statistical properties of each input sample. This dynamic adjustment is achieved by analyzing input data in real-time, extracting key statistical features, and updating the model’s internal parameters—such as weights and biases—accordingly. For instance, in medical image fusion, TTT can analyze the statistical properties of CT and MRI images at test time and dynamically refine the fusion model’s parameters to produce a more accurate and detailed fused image.

\textbf{TTT as a Form of Self-Supervised Learning.}
Moreover, TTT can be considered a form of self-supervised learning applied during inference. Instead of relying on pre-labeled training data, TTT formulates a self-supervised task using the test data itself, allowing the model to learn and adapt to the specific characteristics of the input. This contrasts with traditional supervised learning, where models depend entirely on pre-existing labeled datasets.

By discovering underlying structures and relationships within the test data, TTT enhances its performance on the given task without requiring additional supervision. This makes TTT particularly well suited for applications in medical imaging, where input data often exhibits high variability, such as differences in anatomical structures, imaging artifacts, and disease manifestations. By enabling real-time adaptation, TTT offers a groundbreaking approach to improving the accuracy, efficiency, and reliability of medical image fusion and other AI-driven healthcare applications.

\textbf{The Proposed TTTFusion.}
In the TTTFusion, as shown in Figure \ref{fig:TTTfusion2}, the initial step involves computing feature statistics. Given the feature maps of the MRI ($\mathbf{F}_{MRI}$) and CT ($\mathbf{F}_{CT}$) modalities, where a feature map $\mathbf{F} \in \mathbb{R}^{H\times W\times C}$ with $H$, $W$, and $C$ denoting the height, width, and number of channels, respectively, statistical metrics are derived. The mean $\mu_{c}$ for each channel $c$ is calculated as 
\begin{equation}
\mu_{c} = \frac{1}{H\times W}\sum_{i = 1}^{H}\sum_{j = 1}^{W}F_{ijc},
\end{equation}
and the variance $\sigma^{2}_{c}$ is given by 
\begin{equation}
\sigma^{2}_{c}=\frac{1}{H\times W}\sum_{i = 1}^{H}\sum_{j = 1}^{W}(F_{ijc}-\mu_{c})^{2}.
\end{equation}
These statistics summarize the distribution of features within each modality, facilitating the identification of the most relevant features for accurate medical diagnosis. This process is crucial, as it provides a basis for subsequent operations by capturing the unique characteristics of each modality's features.

Subsequently, feature normalization is indispensable to address the scale disparities among modalities. Z - score normalization, for instance, is applied to transform the feature maps. For a feature map $\mathbf{F}$ with mean $\mu$ and standard deviation $\sigma$, the normalized feature map $\mathbf{\hat{F}}$ is obtained via 
\begin{equation}
\hat{F}_{ijc}=\frac{F_{ijc}-\mu_{c}}{\sigma_{c}+\epsilon},
\end{equation}
with $\epsilon$ being a small positive constant to avoid division by zero. By standardizing the features from different modalities, normalization ensures that no single modality's features dominate the fusion process due to scale differences, creating a balanced environment for information integration.

The kernel calculation and dynamic tuning mechanism are at the core of TTTFusion. Two functions, $f_{MRI}$ and $f_{CT}$, are defined to map the statistical features (mean and variance) of MRI and CT modalities to the weights ($W_1$, $W_2$) and biases ($B_1$, $B_2$) for each modality, expressed as 
\begin{align}
W_1, B_1 &= f_{MRI}(\mu_{MRI},\sigma_{MRI}), \\
W_2, B_2 &= f_{CT}(\mu_{CT},\sigma_{CT}).
\end{align}
The dynamic adjustment of these parameters is driven by the specific statistical characteristics of each input image pair. The weights determine the contribution of each modality's features during fusion, with larger weights indicating greater importance based on statistical analysis. The biases further fine - tune the feature representation, allowing the model to adapt to the diverse distributions of features within each modality. This dynamic adjustment mechanism is essential as it enables the module to optimize the fusion process according to the unique features of the input images, enhancing the utilization of modality - specific information.

Finally, the normalized features from MRI and CT modalities, $\mathbf{\hat{F}}_{MRI}$ and $\mathbf{\hat{F}}_{CT}$, are fused using a weighted combination, resulting in the fused feature map 
\begin{equation}
\mathbf{F}_{fused}=W_1\odot\mathbf{\hat{F}}_{MRI}+W_2\odot\mathbf{\hat{F}}_{CT}+B_1 + B_2,
\end{equation}
where $\odot$ represents element - wise multiplication. The fused feature map encapsulates the most relevant information from both modalities, optimized for the specific input images. Then it is passed to the image reconstruction module for generating the final fused medical image.

\textbf{Multimodal Medical Image Fusion with TTTFusion.} 
Our proposed method integrates the TTTFusion module into the multimodal medical image fusion framework, combining the strengths of traditional feature extraction and reconstruction with the dynamic adaptability of TTT.
During the training phase, the model learns optimal feature representations for each imaging modality, establishing a strong foundation for effective fusion. In the testing phase, the TTTFusion module dynamically adjusts the fusion strategy based on the unique characteristics of the input data. This innovative approach enhances the flexibility and robustness of the model when handling complex medical images, resulting in higher fusion quality and ultimately improving diagnostic accuracy in multimodal medical imaging tasks.

\section{Experiments}

\subsection{Implementation Details}
All experiments were conducted using Python 3.8 and PyTorch 1.13.0. The hardware setup consisted of a desktop computer equipped with a NVIDIA GeForce RTX 4070 Ti GPU and 32GB of total memory.
We utilized two multimodal datasets in our experiments: MRI-CT (184 pairs) and MRI-SPECT (357 pairs) from The Harvard Whole Brain Atlas. For both MRI-CT and MRI-SPECT pairs, we trained the autoencoder for 50 epochs, starting with an initial learning rate of 0.0001, which was decayed cosine-wise to 3e-7. We used a mini-batch size of 4 and the Adam optimizer for training.
To evaluate model performance, we randomly selected 30 image pairs from the MRI-CT dataset and 50 pairs from the MRI-SPECT dataset as a standalone test set. To ensure robustness, we repeated each experiment three times, using different test sets in each run.

\begin{figure}[t]
    \centering
    \includegraphics[width=0.5\textwidth]{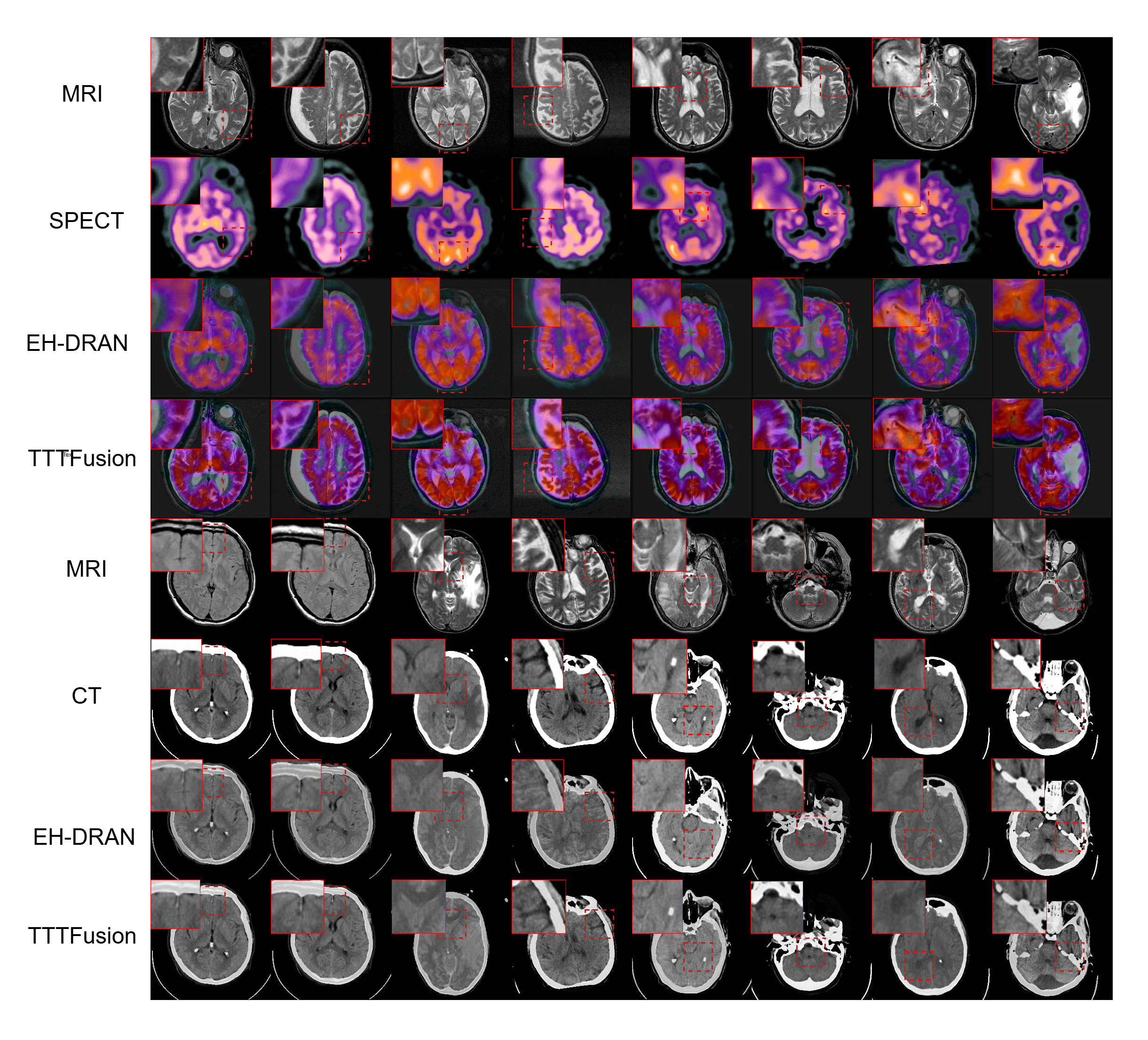}
    \caption{This figure compares the original MRI, SPECT and CT
    images with their fused results by EH-DRAN and TTTFusion. Red boxes aid in the comparison.
}
    \label{fig:compare}
\end{figure}

\begin{table*}[htbp]
    \begin{adjustwidth}{-1in}{-1in} 
        \centering
        \captionsetup{width=0.8\linewidth, justification=centering}
        \caption{COMPARISON BETWEEN DIFFERENT METHODS ON TWO TEST DATASETS. THE RED COLOR  FONT REPRESENTS THE BEST PERFORMANCE AND THE ORANGE COLOR FONT REPRESENTS THE SECOND BEST PERFORMANCE.}
        \label{tab:results}
        \begin{tabular}{lcccccc}
            \hline
            Dataset & Method & PSNR & SSIM & FMI & FSIM & EN \\
            \hline
            \multirow{7}{*}{MRI-CT} 
                & IFCNN\cite{Zhang2020Ifcnn} & $15.594 \pm 0.112$ & $0.700 \pm 0.015$ & $0.870 \pm 0.012$ & $0.801 \pm 0.001$ & $8.968 \pm 0.227$ \\
                & MSRPAN\cite{Fu2021Multiscale}& $14.790 \pm 0.233$ & $0.749 \pm 0.003$ & $0.744 \pm 0.001$ & $0.804 \pm 0.001$ & $7.773 \pm 0.273$ \\
                & MSDRA\cite{Li2022Multiscale}& $15.308 \pm 0.437$ & $0.742 \pm 0.037$ & $0.872 \pm 0.002$ & $0.788 \pm 0.005$ & $9.554 \pm 0.767$ \\
                & SwinFusion\cite{Ma2022Swinfusion} & $14.962 \pm 0.173$ & \textcolor{red}{$0.768 \pm 0.007$ }& $0.882 \pm 0.002$ & $0.810 \pm 0.001$ & $8.445 \pm 0.078$ \\
                & MRSCFusion\cite{Xie2023Mrscfusion} & $14.476 \pm 0.205$ & $0.713 \pm 0.012$ & $0.877 \pm 0.006$ & $0.791 \pm 0.010$ & $7.544 \pm 0.232$ \\
                & EH-DRAN\cite{zhou2024edge}& \textcolor{orange}{$16.830 \pm 0.490$} & $0.753 \pm 0.007$ & \textcolor{orange}{$0.883 \pm 0.005$} & \textcolor{orange}{$0.820 \pm 0.003$} &\textcolor{orange}{ $10.727 \pm 0.531$} \\
                & \textbf{TTTFusion (Ours)} & \textcolor{red}{$17.630 \pm 0.127$} & \textcolor{orange}{$0.763 \pm 0.011$} & \textcolor{red}{$0.895 \pm 0.006$} & \textcolor{red}{$0.831 \pm 0.002$} & \textcolor{red}{$12.081 \pm 1.780$} \\
            \hline
            \multirow{7}{*}{MRI-SPECT} 
                & IFCNN\cite{Zhang2020Ifcnn}& $19.728 \pm 0.228$ & $0.721 \pm 0.025$ & $0.846 \pm 0.062$ & $0.783 \pm 0.027$ & $10.167 \pm 0.429$ \\
                & MSRPAN\cite{Fu2021Multiscale} & $19.174 \pm 0.046$ & $0.732 \pm 0.002$ & $0.838 \pm 0.003$ & $0.793 \pm 0.002$ & $9.737 \pm 0.202$ \\
                & MSDRA\cite{Li2022Multiscale} & $19.662 \pm 0.165$ & $0.725 \pm 0.003$ & $0.839 \pm 0.003$ & $0.794 \pm 0.003$ & $10.784 \pm 0.447$ \\
                & SwinFusion\cite{Ma2022Swinfusion} & $17.557 \pm 0.021$ & $0.728 \pm 0.004$ & $0.808 \pm 0.007$ & $0.819 \pm 0.011$ & $13.066 \pm 0.428$ \\
                & MRSCFusion\cite{Xie2023Mrscfusion}& $18.412 \pm 0.211$ & $0.734 \pm 0.012$ & $0.827 \pm 0.009$ & $0.814 \pm 0.006$ & $9.870 \pm 0.600$ \\
                & EH-DRAN\cite{zhou2024edge} & \textcolor{orange}{$21.455 \pm 0.071$} & \textcolor{orange}{$0.736 \pm 0.002$} & \textcolor{orange}{$0.876 \pm 0.004$} & \textcolor{orange}{$0.843 \pm 0.003$} & \textcolor{orange}{$11.970 \pm 0.538$} \\
                & \textbf{TTTFusion (Ours)} & \textcolor{red}{$22.178 \pm 0.203$} & \textcolor{red}{$0.746 \pm 0.007$} & \textcolor{red}{$0.886 \pm 0.003$} & \textcolor{red}{$0.859 \pm 0.002$} & \textcolor{red}{$12.434 \pm 1.681$} \\
            \hline
        \end{tabular}
        \label{table:1}
    \end{adjustwidth}
\end{table*}

\begin{table*}[htbp]
    \begin{adjustwidth}{-1.5in}{-1.5in} 
        \centering
        \captionsetup{width=0.8\linewidth, justification=centering}
        \caption{COMPARISON BETWEEN DIFFERENT FUSION STRATEGIES ON TWO TEST DATASETS}
        \label{tab:fusion_results}
        \begin{tabular}{lcccccc}
            \hline
            Dataset & Fusion Strategy & PSNR & SSIM & FMI & FSIM & EN \\
            \hline
            \multirow{6}{*}{MRI - CT} 
                & FER\cite{Fu2021Multiscale}& $14.718\pm0.549$ & $0.743\pm0.005$ & $0.874\pm0.002$ & $0.798\pm0.008$ & $8.675\pm0.182$ \\
                & FLIN\cite{Li2022Multiscale}& $15.620\pm0.279$ & $0.737\pm0.002$ & $0.878\pm0.003$ & $0.804\pm0.010$ & $9.001\pm0.583$ \\
                & SFNN - mean\cite{zhou2024edge}& $15.631\pm0.290$ & $0.736\pm0.003$ & $0.877\pm0.002$ & $0.810\pm0.010$ & $9.013\pm0.351$ \\
                & SFNN - max\cite{zhou2024edge}& \textcolor{orange}{$16.830\pm0.490$} & \textcolor{orange}{$0.753\pm0.007$} & \textcolor{orange}{$0.883\pm0.005$} & \textcolor{orange}{$0.820\pm0.003$} & \textcolor{orange}{$10.727\pm0.531$} \\
                & SFNN - sum\cite{zhou2024edge} & $15.590\pm0.370$ & $0.735\pm0.005$ & $0.877\pm0.004$ & $0.810\pm0.007$ & $8.303\pm0.618$ \\
                & \textbf{TTTFusion (Ours)} & \textcolor{red}{$17.630\pm0.127$} & \textcolor{red}{$0.763\pm0.011$} & \textcolor{red}{$0.895\pm0.006$} & \textcolor{red}{$0.831\pm0.002$} & \textcolor{red}{$12.081\pm1.780$} \\
            \hline
            \multirow{6}{*}{MRI - SPECT} 
                & FER\cite{Fu2021Multiscale} & $19.635\pm0.039$ & $0.832\pm0.004$ & $0.832\pm0.005$ & $0.796\pm0.003$ & $9.751\pm0.112$ \\
                & FLIN\cite{Li2022Multiscale}& $20.337\pm0.058$ & $0.833\pm0.002$ & $0.842\pm0.006$& $0.800\pm0.010$ & $10.562\pm0.482$\\
                & SFNN - mean\cite{zhou2024edge} & $20.337\pm0.043$ & $0.734\pm0.003$& $0.841\pm0.003$ & $0.836\pm0.006$& $10.115\pm0.574$\\
                & SFNN - max\cite{zhou2024edge} & \textcolor{orange}{$21.455\pm0.071$} & \textcolor{orange}{$0.736\pm0.002$} & \textcolor{orange}{$0.876\pm0.004$} & \textcolor{orange}{$0.843\pm0.003$} & \textcolor{orange}{$11.970\pm0.538$} \\
                & SFNN - sum\cite{zhou2024edge}& $20.336\pm0.136$ & $0.734\pm0.002$ & $0.842\pm0.003$ & $0.838\pm0.010$ & $10.937\pm0.451$ \\
                & \textbf{TTTFusion (Ours)} & \textcolor{red}{$22.178\pm0.203$} & \textcolor{red}{$0.746\pm0.007$} & \textcolor{red}{$0.886\pm0.003$} & \textcolor{red}{$0.859\pm0.002$} & \textcolor{red}{$12.434\pm1.681$} \\
            \hline
        \end{tabular}
            \label{table:2}
    \end{adjustwidth}
\end{table*}

\subsection{Results and Discussions}

In this section, we present the experimental results of our proposed TTTFusion strategy for image fusion tasks, comparing it with several state-of-the-art methods, including IFCNN \cite{Zhang2020Ifcnn}, MSRPAN \cite{Fu2021Multiscale}, MSDRA \cite{Li2022Multiscale}, SwinFusion \cite{Ma2022Swinfusion}, MRSCFusion \cite{Xie2023Mrscfusion}, and EH-DRAN \cite{zhou2024edge}. We focus on the MRI-CT and MRI-SPECT datasets, analyzing the performance of TTTFusion through both quantitative and qualitative evaluations.

As shown in Figure \ref{fig:compare}, the primary objective of the MRI-CT fusion task is to preserve detailed tissue information from MRI scans while maintaining bone structures and other dense features from CT scans. Our TTTFusion method demonstrates significant improvements over baseline methods, particularly in retaining fine-grained details from both modalities. TTTFusion achieves superior delineation of brain contours, enhanced edge contrast, and better tissue preservation compared to EH-DRAN, SwinFusion, and MRSCFusion. Specifically, TTTFusion enhances the clarity of soft tissue structures from MRI (e.g., brain contours) while preserving high-density features from CT (e.g., bone structures) with exceptional edge contrast. In the MRI-SPECT fusion task, the goal is to preserve both functional and morphological details. SPECT images provide critical information on metabolic activity and blood flow, while MRI images offer anatomical structures. TTTFusion outperforms other methods, including EH-DRAN, by retaining more detailed functional and morphological features from both MRI and SPECT images. The fusion results exhibit better preservation of color and texture information from SPECT images and structural details from MRI scans, resulting in fused images with high contrast and structural clarity.

We evaluated the performance of TTTFusion using several widely adopted fusion metrics: Peak Signal-to-Noise Ratio (PSNR), Structural Similarity Index (SSIM), Feature Mutual Information (FMI), Feature Structural Similarity (FSIM), and Information Entropy (EN). The results, summarized in Table \ref{table:1}, demonstrate that TTTFusion consistently outperforms all baseline methods in these metrics. In the MRI-CT fusion task, TTTFusion achieves significantly higher FMI and FSIM scores compared to EH-DRAN, SwinFusion, and MRSCFusion, indicating better preservation of structural integrity and fine-grained features. The improvements in PSNR and EN further confirm the enhanced quality and reduced information loss in the fused images generated by TTTFusion. In the MRI-SPECT fusion task, TTTFusion again shows superior performance, achieving the highest scores in PSNR, SSIM, FMI, and FSIM. This highlights its ability to effectively preserve both functional and morphological details of input modalities. The improved performance is particularly evident in the retention of fine texture details from both MRI and SPECT images, which is critical for accurate medical image analysis.

To evaluate the effectiveness of TTTFusion, we compared it with several other fusion strategies. The results, presented in Table \ref{table:2}, show that TTTFusion consistently outperforms these strategies in all metrics in both the MRI-CT and MRI-SPECT fusion tasks. TTTFusion provides superior visual quality, maintaining sharper delineation of soft tissue structures from MRI images, and better preservation of dense structures from CT and SPECT images. The contrast between the edges and the inner tissues in the fused images is notably improved, offering a more natural and clinically relevant representation of the anatomical and functional details. Although methods like FER and FLIN are effective, they struggle to achieve the same level of detail preservation and edge contrast. For instance, FER tends to blur fine details, and FLIN sometimes introduces intensity distortions, particularly in regions where high-density structures from CT images should be preserved.

The superior performance of TTTFusion can be attributed to its novel fusion mechanism, which leverages multi-scale feature extraction and a parameter-free fusion strategy. The ability to dynamically adapt to input data during the test phase allows TTTFusion to achieve state-of-the-art results in both MRI-CT and MRI-SPECT fusion tasks. Quantitative and qualitative improvements highlight the potential of TTTFusion for real-time medical applications, such as surgical navigation, where accurate and efficient image fusion is critical. In conclusion, TTTFusion sets a new benchmark for medical image fusion, addressing key challenges in detail preservation, edge contrast, and real-time adaptability. Its performance improvements over existing methods underscore its potential for advancing medical imaging and analysis.

\section{Conclusion}



This paper presents TTTFusion, a Test-Time Training-based multimodal medical image fusion strategy, and applies it to surgical robot systems. By dynamically adjusting model parameters during inference, TTTFusion significantly enhances fusion quality, particularly in fine-grained feature extraction and edge information preservation.
Experimental results demonstrate that, compared to traditional fusion methods, TTTFusion not only improves image fusion accuracy but also effectively addresses modality differences, especially in preserving the details of complex lesions. This approach introduces a new technological pathway for real-time image processing in surgical robots, offering promising applications in the field.

For future work, we plan to further optimize the TTTFusion strategy and explore its use in more complex clinical settings, particularly for the fusion of diverse medical imaging modalities. Additionally, we aim to integrate TTTFusion into surgical robot systems to enhance operational efficiency and precision in real-world surgeries.

\end{document}